\pdfoutput=1

\documentclass[11pt]{article}

\usepackage[preprint]{acl}

\usepackage{times}
\usepackage{latexsym}

\usepackage[T1]{fontenc}

\usepackage[utf8]{inputenc}

\usepackage{microtype}

\usepackage{inconsolata}

\usepackage{algorithm}
\usepackage{algorithmic}
\usepackage{amsmath}
\usepackage{bbm}
\usepackage{multicol}
\usepackage{hyperref}
\usepackage[nameinlink]{cleveref}

\usepackage{amssymb}
\usepackage{pifont}
\newcommand{\cmark}{\text{\ding{51}}}%
\newcommand{\xmark}{\text{\ding{55}}}%

\usepackage{color}
\usepackage{tabularray}

\usepackage{newunicodechar}

\usepackage{caption}
\usepackage{subcaption}

\usepackage{blindtext}
\usepackage{graphicx}
\usepackage{makecell}

\usepackage{booktabs}
\usepackage[normalem]{ulem}
\useunder{\uline}{\ul}{}

\usepackage{lipsum}
\usepackage{tikz}
\usepackage{fancyvrb}
\usepackage{listings}
\usepackage{enumitem}
\usepackage{tcolorbox}
\usepackage{multicol}
\usepackage{siunitx}
\usepackage{xpatch}
\usepackage{arydshln} 

\usepackage{multirow}  

\title{ImplicitAVE: An Open-Source Dataset and Multimodal LLMs Benchmark for Implicit Attribute Value Extraction}

\author{Henry Peng Zou$^{1}$, Vinay Samuel$^{2}$, Yue Zhou$^{1}$, Weizhi Zhang$^{1}$, \\ \bf Liancheng Fang$^{1}$, Zihe Song$^{1}$, Philip S. Yu$^{1}$, Cornelia Caragea$^{1}$
\\  $^{1}$University of Illinois Chicago \quad $^2$Carnegie Mellon University \\
{\color{blue}\texttt{\{pzou3,yzhou232,wzhan42,lfang87,zsong29,psyu,cornelia\}@uic.edu}} \\
{\color{blue}\texttt{vsamuel@andrew.cmu.edu}}
}

\begin{document}
\maketitle
\begin{abstract}
Existing datasets for attribute value extraction (AVE) predominantly focus on explicit attribute values while neglecting the implicit ones, lack product images, are often not publicly available, and lack an in-depth human inspection across diverse domains. To address these limitations, we present ImplicitAVE, the first, publicly available multimodal dataset for implicit attribute value extraction. ImplicitAVE, sourced from the MAVE dataset, is carefully curated and expanded to include implicit AVE and multimodality, resulting in a refined dataset of 68k training and 1.6k testing data across five domains. We also explore the application of multimodal large language models (MLLMs) to implicit AVE, establishing a comprehensive benchmark for MLLMs on the ImplicitAVE dataset. Six recent MLLMs with eleven variants are evaluated across diverse settings, revealing that implicit value extraction remains a challenging task for MLLMs. The contributions of this work include the development and release of ImplicitAVE, and the exploration and benchmarking of various MLLMs for implicit AVE, providing valuable insights and potential future research directions. Dataset and code are available at 
\textcolor{blue}{\url{https://github.com/HenryPengZou/ImplicitAVE}}.
\end{abstract}


\section{Introduction}

\begin{figure}[!t]
    \centering
    \includegraphics[width=\columnwidth]{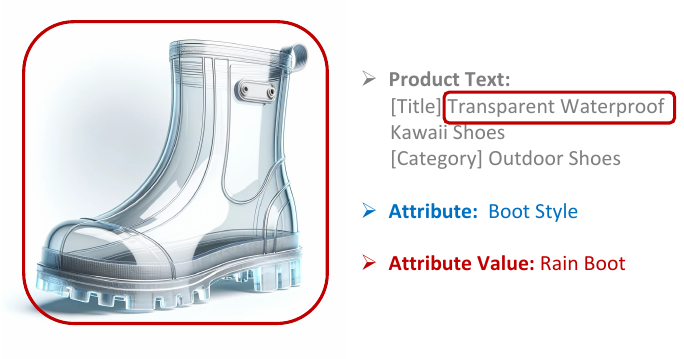} 
    \caption{An example of implicit attribute value. The attribute value {\color{red!70!black}``Rain Boot''} is not mentioned explicitly in the product text, but can be inferred from text context, product image, or prior knowledge.}
    \label{fig:implicit_values_example}
     \vspace{-4mm}
\end{figure}

Attribute Value Extraction (AVE) identifies the value of product attributes from the product information, which is critical in e-commerce for product representation, recommendation, and categorization \cite{mave, aveqa, khandelwal2023large, yang2023mixpave, fang2024llm}. The attribute values can be categorized into two types: (1) \emph{Explicit} values can be directly found as a segment in the product text \cite{mave, aveqa}, while (2) \emph{Implicit} values are never mentioned in the text and can only be inferred from the product image, contextual clues, or prior knowledge \cite{zhang-etal-2023-pay}. Consider the example in Figure~\ref{fig:implicit_values_example}. The \texttt{value} ``rain boot'' of the \texttt{attribute} ``boot style'' is implicit since it is not explicitly stated in the product text but can be inferred from its image or context from keywords such as ``transparent'' and ``waterproof.''


\begin{table*}[!t]
\centering
\resizebox{\textwidth}{!}{%
\begin{tabular}{@{}lcccccc@{}}
\toprule

\textbf{Dataset} & \textbf{Implicit Values} & \textbf{Multimodality} & \textbf{Publicly Available} & \textbf{Human Annotation} & \textbf{Multiple Domains} & \textbf{Language} \\ \midrule
OpenTag \cite{opentag} & \textcolor{purple}{\xmark} & \textcolor{purple}{\xmark} & \textcolor{purple}{\xmark} & \textcolor{teal}{\cmark} & \textcolor{teal}{\cmark} & English \\
AE-110K \cite{xu-etal-2019-scaling} & \textcolor{purple}{\xmark} & \textcolor{purple}{\xmark} & \textcolor{teal}{\cmark} & \textcolor{purple}{\xmark} & \textcolor{teal}{\cmark} & Chinese \\
MEPAVE \cite{zhu-etal-2020-multimodal} & \textcolor{purple}{\xmark} & \textcolor{teal}{\cmark} & \textcolor{teal}{\cmark} & \textcolor{teal}{\cmark} & \textcolor{teal}{\cmark} & Chinese \\
AdaTag \cite{yan-etal-2021-adatag} & \textcolor{purple}{\xmark} & \textcolor{purple}{\xmark} & \textcolor{purple}{\xmark} & \textcolor{teal}{\cmark} & \textcolor{purple}{\xmark} & English \\
MAVE \cite{mave} & \textcolor{purple}{\xmark} & \textcolor{purple}{\xmark} & \textcolor{teal}{\cmark} & \textcolor{purple}{\xmark} & \textcolor{teal}{\cmark} & English \\
DESIRE \cite{zhang-etal-2023-pay} & \textcolor{teal}{\cmark} & \textcolor{teal}{\cmark} & \textcolor{purple}{\xmark} & \textcolor{teal}{\cmark} & \textcolor{purple}{\xmark} & Chinese \\ \midrule
ImplicitAVE (Ours) & \textcolor{teal}{\cmark} & \textcolor{teal}{\cmark} & \textcolor{teal}{\cmark} & \textcolor{teal}{\cmark} & \textcolor{teal}{\cmark} & English \\ \bottomrule
\end{tabular}%
}
\caption{Comparison of existing AVE datasets. While several \textbf{\textit{explicit}} AVE datasets exist, \textbf{\textit{implicit}} AVE is much more challenging and under-explored. Our work introduces the first open-source dataset that is expressly designed to address the task of implicit AVE. Our dataset is considerably different from DESIRE, as detailed in Appendix \ref{sec:dif_desire}.}
\label{tab:dataset_compare}
\end{table*}

Nonetheless, existing datasets for attribute value extraction exhibit several key limitations: (1) They predominantly focus on explicit attribute values, neglecting implicit attribute values \cite{opentag, aveqa}, which are more challenging and commonly encountered in real-world scenarios; (2) Many datasets lack product images \cite{yan-etal-2021-adatag, mave}, limiting their applicability in multimodal contexts; (3) The limited number of publicly available datasets lack human inspection and cover only a few domains, resulting in inaccurate and restricted benchmarks \cite{xu-etal-2019-scaling, zhang-etal-2023-pay}. Table~\ref{tab:dataset_compare} compares these aspects for various AVE datasets.

To address these issues, we present ImplicitAVE, the first publicly available multimodal dataset for implicit attribute value extraction. We initially sourced product text data from the MAVE dataset \cite{mave} and then curated the data by eliminating unhelpful attributes and redundant or irrelevant values. Subsequently, we transformed and expanded the dataset to include implicit attribute value extraction and multimodality and finally validated the test set annotations through two rounds of human inspection. This yields a more refined and quality-improved dataset of 68k training and 1.6k testing data spanning five diverse domains with 25 attributes and corresponding attribute values suitable for implicit attribute value extraction. Detailed statistics of our dataset are shown in Tables~\ref{tab:dataset_statistics_domain}, \ref{tab:dataset_statistics_attribute_level}.

Given the cutting-edge performance of Multimodal Large Language Models (MLLMs) \cite{BLIP-2, LLaVA,Liu2023ImprovedBW,bai2023qwen,ye2023mplug,luo2023cheap} and the absence of previous exploration of their application to implicit attribute value extraction, we establish a comprehensive benchmark for MLLMs on our ImplicitAVE dataset. We cover six recent MLLMs with 11 variants and compare them with the fine-tuned previous SOTA method. We evaluate their performance across diverse settings, including full/few-shot and zero-shot scenarios, domain-level and attribute-level performance, and single/multi-modality performance. We find that implicit value extraction remains a challenging task for open-source MLLMs despite their effective capabilities. 

Our contributions are summarized as follows: (1) The development and release of ImplicitAVE, the first open-source multimodal dataset for implicit AVE; (2) The exploration and benchmarking of various MLLMs for implicit attribute value extraction across diverse settings, revealing intriguing insights and potential future research directions.



\begin{figure*}[!th]
    \centering
    \includegraphics[width=\textwidth]{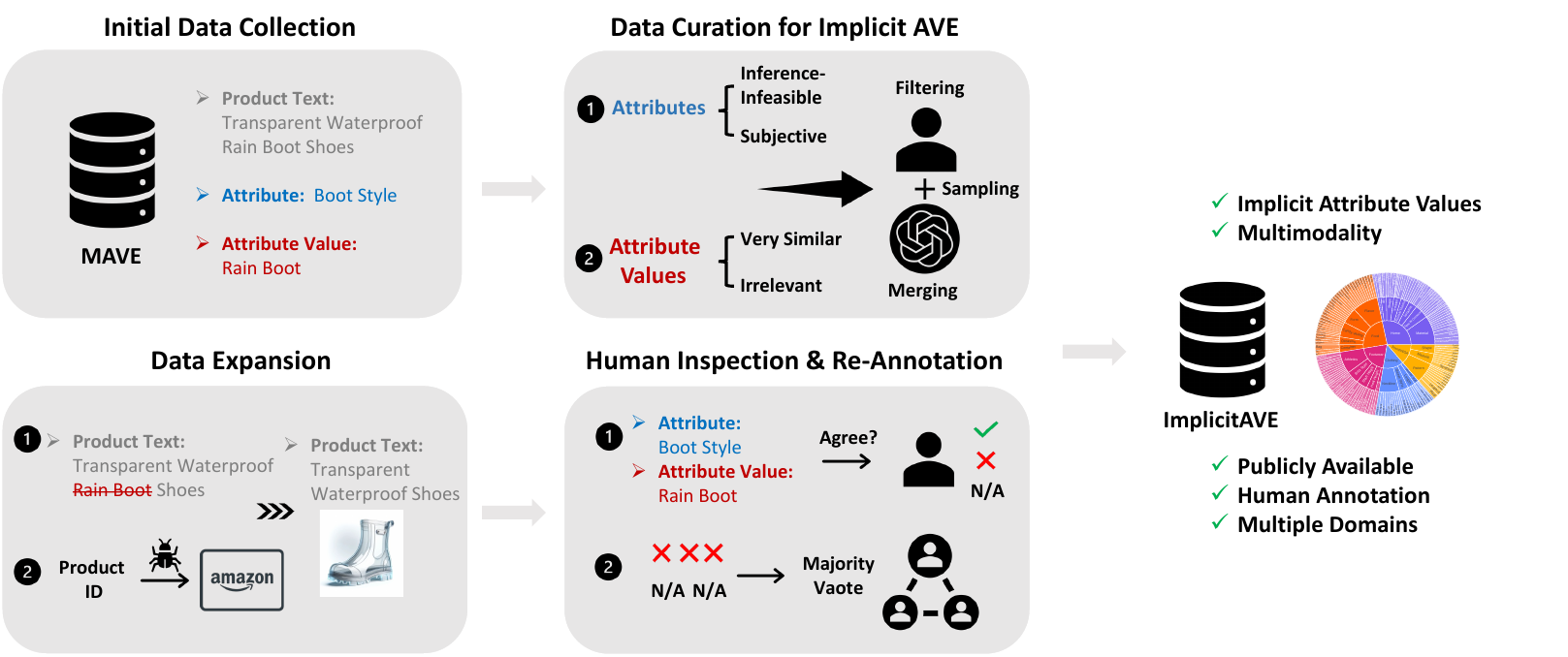} 
    \caption{Steps for constructing our ImplicitAVE dataset. A detailed explanation is provided in Section \ref{sec:dataset_construction}.}
    \label{fig:pipeline}
\end{figure*}


\begin{table*}[!bht]
\centering
\resizebox{\textwidth}{!}{%
\begin{tabular}{@{}lccccl@{}}
\toprule
\textbf{Domain} & \textbf{\# Train} & \textbf{\# Eval} & \textbf{\# Values} & \textbf{\# Attributes} & \multicolumn{1}{c}{\textbf{Attributes}} \\ \midrule
Clothing & 18868 & 226 & 23 & 4 & {[}'Sleeve Style', 'Neckline', 'Length', 'Shoulder Style'{]} \\
Footwear & 21442 & 317 & 29 & 5 & {[}'Shaft Height', 'Athletic Shoe Style', 'Boot Style', 'Heel Height', 'Toe Style'{]} \\
Jewelry\&GA & 13061 & 220 & 20 & 3 & {[}'Pattern', 'Material', 'Shape'{]} \\
Food & 3617 & 390 & 41 & 5 & {[}'Form', 'Candy Variety', 'Container', 'Occasion', 'Flavor'{]} \\
Home Product & 11616 & 457 & 45 & 8 & {[}'Season', 'Material', 'Location', 'Animal Theme', 'Special Occasion', 'Size', 'Attachment Method', 'Shape'{]} \\
All & 68604 & 1610 & 158 & 25 & - \\ \bottomrule
\end{tabular}%
}
\caption{Domain-level dataset statistics.}
\label{tab:dataset_statistics_domain}
\end{table*}


\section{Dataset Construction}
\label{sec:dataset_construction}






We outline our approach to constructing the first open-source multimodal implicit attribute value extraction dataset, ImplicitAVE. The dataset construction pipeline is illustrated in Figure \ref{fig:pipeline}. It contains four steps: data collection, curation, expansion, and validation. Next, we explain them in detail.

\subsection{Initial Data Collection}
Initially, we sourced product text information, including titles, categories, and corresponding attribute-value annotations, from the publicly available MAVE dataset \cite{mave}, comprising 2.2 million products spanning diverse e-commerce domains. Despite its extensive coverage, the MAVE dataset exhibits several significant \emph{limitations}, making it unsuitable for implicit AVE: (1) It contains inappropriate attributes and values that are not facilitative to implicit AVE tasks (see Step 2); (2) It is designed solely for explicit attribute-value extraction; (3) It solely comprises textual information and lacks multimodal data sources; (4) Annotations within the MAVE dataset are machine-generated and lack human inspection, resulting in notable inaccuracies. 

\subsection{Data Curation for Implicit AVE}
We further refine the sourced data by removing unhelpful attributes and redundant or irrelevant values for Implicit AVE. Concretely: \ding{182} \textbf{Removing Inference-Infeasible Attributes}. We manually inspect and remove attributes where the specific values are almost impossible to infer if the values are not mentioned explicitly in the text, such as display resolution, storage capacity, and battery life; \ding{183} \textbf{Removing Subjective Attributes}. The attributes that are rather subjective and ambiguous, such as the degree of comfort and product quality, are also removed; \ding{184} \textbf{Value Merging and Cleaning}. Attribute values with similar semantic meanings are consolidated. This includes unifying variations in grammar forms (e.g., Short-Sleeve, Short sleeves, short sleeved for the attribute Sleeve Style), eliminating extraneous words (e.g., running and running shoes), and merging synonyms (e.g., floral and flower, leopard and cheetah, crew neck and round neck, plaid and tartan, etc.) In addition, we notice some values are irrelevant to their parent attributes (e.g., the value ``Clear Stamps'' of the attribute ``Material of Artwork''), so these values are removed as well. The value merging and cleaning is achieved collaboratively by lexicon-based scripts, prompting with GPT-4, and human inspection.

This curation results in a more refined and quality-improved dataset with 25 attributes and corresponding attribute values spanning five domains suitable for implicit attribute value extraction. We randomly sample up to 1000 instances per attribute value to limit the dataset size. The selected domains and attributes in ImplicitAVE are shown in Table \ref{tab:dataset_statistics_domain}.

\begin{table*}[!thb]
\centering
\resizebox{\textwidth}{!}{%
\begin{tabular}{@{}llcccl@{}}
\toprule
\textbf{Domains} & \textbf{Attributes} & \textbf{\# Train} & \textbf{\# Eval} & \textbf{\# Values} & \textbf{Attribute Values} \\ \midrule
 & Sleeve Style & 3957 & 50 & 5 & {[}'Short Sleeve', 'Long Sleeve', '3/4 Sleeve', 'Sleeveless', 'Strappy'{]} \\
 & Neckline & 8141 & 110 & 11 & {[}'Crew Neck', 'V-Neck', 'Henley', 'Polo', 'Scoop Neck', 'Strapless', 'Button Down', ...{]} \\
Clothing & Length & 4937 & 40 & 4 & {[}'Mini/Short', 'Midi', 'Long Dress/Gown', 'Capri'{]} \\
 & Shoulder Style & 1833 & 26 & 3 & {[}'One Shoulder', 'Off Shoulder', 'Cold Shoulder'{]} \\
 \midrule
 & Shaft Height & 4546 & 60 & 5 & {[}'Ankle Boot', 'Bootie', 'Knee High', 'Mid Calf', 'Over The Knee'{]} \\
 & Athletic Shoe Style & 8165 & 119 & 12 & {[}'Hiking Boot', 'Soccer', 'Golf', 'Running Shoe', 'Basketball', 'Tennis', 'Walking', ...{]} \\
Footwear & Boot Style & 5145 & 68 & 6 & {[}'Western/Cowboy', 'Chelsea', 'Combat', 'Snow Boots', 'Motorcycle', 'Rain Boots'{]} \\
 & Heel Height & 2457 & 50 & 4 & {[}'High Heel', 'Flat', 'Mid Heel', 'Low Heel'{]} \\
 & Toe Style & 1129 & 20 & 2 & {[}'Round Toe', 'Pointed Toe'{]} \\
 \midrule
 & Pattern & 8418 & 111 & 10 & {[}'Floral', 'Camouflage', 'Plaid', 'Leopard', 'Stripe', 'Paisley', 'Polka Dot', 'Argyle', ...{]} \\
Jewelry\&GA & Material & 2390 & 59 & 5 & {[}'Leather', 'Canvas', 'Synthetic', 'Wooden', 'Metal'{]} \\
 & Shape & 2253 & 50 & 5 & {[}'Heart', 'Cross', 'Round', 'Oval', 'Crucifix'{]} \\
 \midrule
 & Form & 1423 & 86 & 9 & {[}'Bags/Packets', 'Powder', 'Teabags', 'Rub', 'Bottles', 'Soup Mix', 'Flakes', 'Sticks', 'Sliced'{]} \\
 & Candy Variety & 798 & 82 & 9 & {[}'Gummy/Chewy', 'Gum', 'Hard Candy', 'Mints', 'Licorice', 'Jelly Beans', 'Mint', 'Lollipop'{]} \\
Food & Container & 563 & 40 & 4 & {[}'Bag', 'Box', 'Tin', 'Case'{]} \\
 & Occasion & 148 & 43 & 5 & {[}'Easter', 'Other Holiday', "Valentine's", 'Halloween', 'Christmas'{]} \\
 & Flavor & 685 & 139 & 14 & {[}'Vanilla', 'Salted', 'Butter', 'Chocolate', 'Original', 'Strawberry', 'Habanero', 'Caramel', ...{]} \\
 \midrule
 & Season & 215 & 40 & 5 & {[}'All Seasons', 'Autumn', 'Spring', 'Summer', 'Winter'{]} \\
 & Material & 7523 & 158 & 13 & {[}'Metal', 'Ceramic/Melamine', 'Fabric', 'Bamboo', 'Silicone', 'Wood', 'Plastic', 'Glass', ...{]} \\
 & Location & 50 & 47 & 4 & {[}'Bedroom', 'Kitchen', 'Outdoor', 'Bathroom'{]} \\
 & Animal Theme & 134 & 46 & 5 & {[}'Cat', 'Dog', 'Owl', 'Bird'{]} \\
Home & Special Occasion & 1002 & 76 & 8 & {[}'Christmas', 'Halloween', 'Wedding', 'Birthday', 'Graduation', 'Patriotic', 'Easter', ...{]} \\
 & Size & 655 & 30 & 4 & {[}'Queen', 'King', 'Full', 'Twin'{]} \\
 & Attachment Method & 441 & 20 & 2 & {[}'Grommet', 'Rod Pocket'{]} \\
 & Shape & 1596 & 40 & 4 & {[}'Square', 'Rectangular', 'Oval', 'Round'{]} \\
 \midrule
All & - & 68604 & 1610 & 158 & - \\ \bottomrule
\end{tabular}%
}
\caption{Attribute-level dataset statistics. The detailed ontology of our data and examples of products in different domains, with different attributes and values are provided in Appendix \ref{sec:appendix_domain_attribute_value}.}
\label{tab:dataset_statistics_attribute_level}
\vspace{-3mm}
\end{table*}

\subsection{Data Expansion}
To extend the data for implicit attribute value extraction and multimodality, we perform the following processing steps: \ding{182} \textbf{Implicit Value Creation}. We remove all explicit attribute value mentions from the input text for its corresponding attribute for each data point. As a result, attribute values in these data can only be inferred from the product images, indirect text context, or prior knowledge. That is, these values become implicit attribute values given the modified inputs. We then drop instances with the same product ID or image to prevent potential information leakage across instances based on the same product. \ding{183} \textbf{Multimodality Creation}. We systematically collect product images from the Amazon website using the product identification number and thus expand our dataset with multimodal information. 

\subsection{Human Inspection \& Re-Annotation}
Through manual inspection, we observed that the original attribute-value annotations from MAVE contain noticeable errors. This is because they were annotated by ensembling predictions from five variations of AVEQA models~\cite{aveqa}\footnote{AVEQA \cite{aveqa} is a question-answering model that regards each query attribute as a question and determines the answer span that matches the attribute value within the product text information.} without human inspection. To rectify incorrect annotations and ensure a high-quality test set for implicit attribute value extraction and MLLMs evaluation, we engage five Ph.D. students to manually inspect and re-annotate our evaluation set. 


This process first involves sampling ten instances per attribute value from the constructed dataset, resulting in 1,676 instances. The human inspection and re-annotation process then unfold in \emph{two} rounds:
In the first round, annotators assess each instance's product image, text contexts, and relevant attributes to determine the correctness of the original attribute value annotation. If annotators think the original annotation is incorrect, they select the best attribute value from the corresponding value list (of that attribute) or mark `N/A' if the annotator believes no suitable value is provided or multiple values are suitable. Additionally, annotators can suggest improvements such as merging, removing, adding, or replacing attribute values. Of the total instances, 1,448 original annotations are correct, 172 are incorrect, and 56 are marked as `N/A,' yielding an agreement rate of 86.4\%. Ten, one, one, and one attribute values are suggested for merging, removing, adding, and replacing, respectively. Instances with disagreed annotations are subject to a second round of inspection and re-annotation, wherein three well-trained annotators participate, and a majority vote determines the final annotation for each instance.

\subsection{Dataset Statistics}

\begin{figure}[t]
     \centering
     \begin{subfigure}[b]{0.8\columnwidth}
         \centering
         \includegraphics[width=\textwidth]{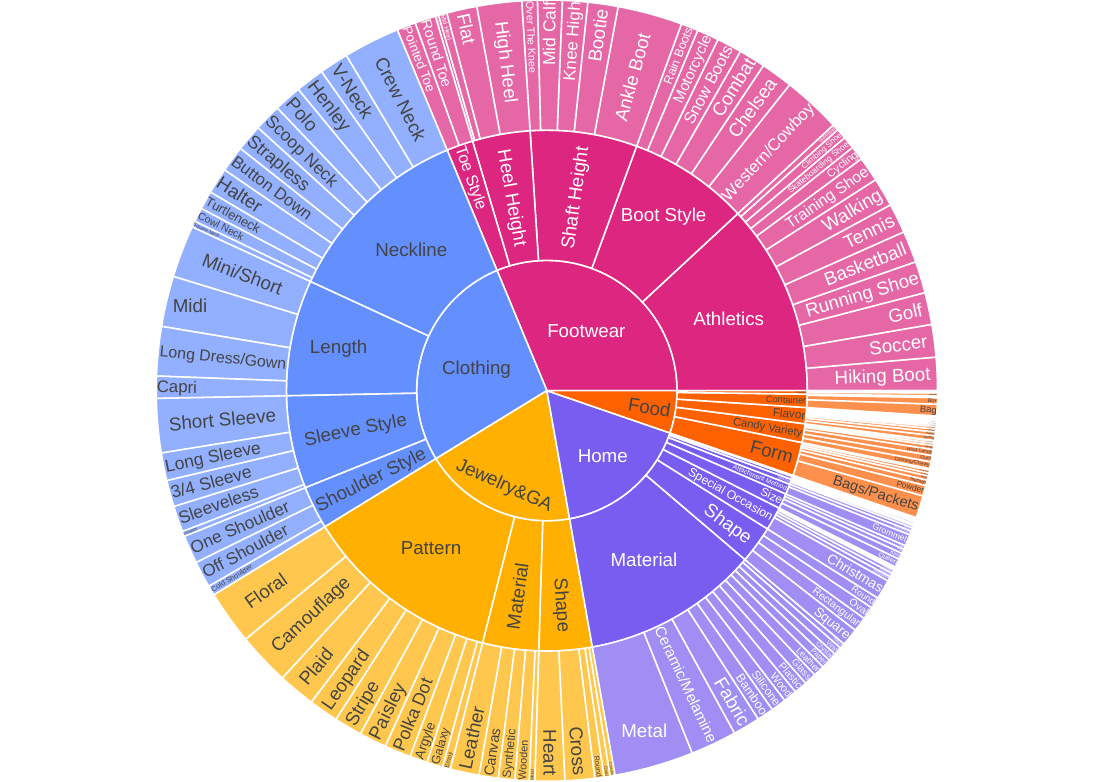}
         \caption{Training Set}
         \vspace{1pt}
         \label{fig:visual_data_distribution_trainining}
     \end{subfigure}
     \hfill
     \begin{subfigure}[b]{0.8\columnwidth}
         \centering
         \includegraphics[width=\textwidth]{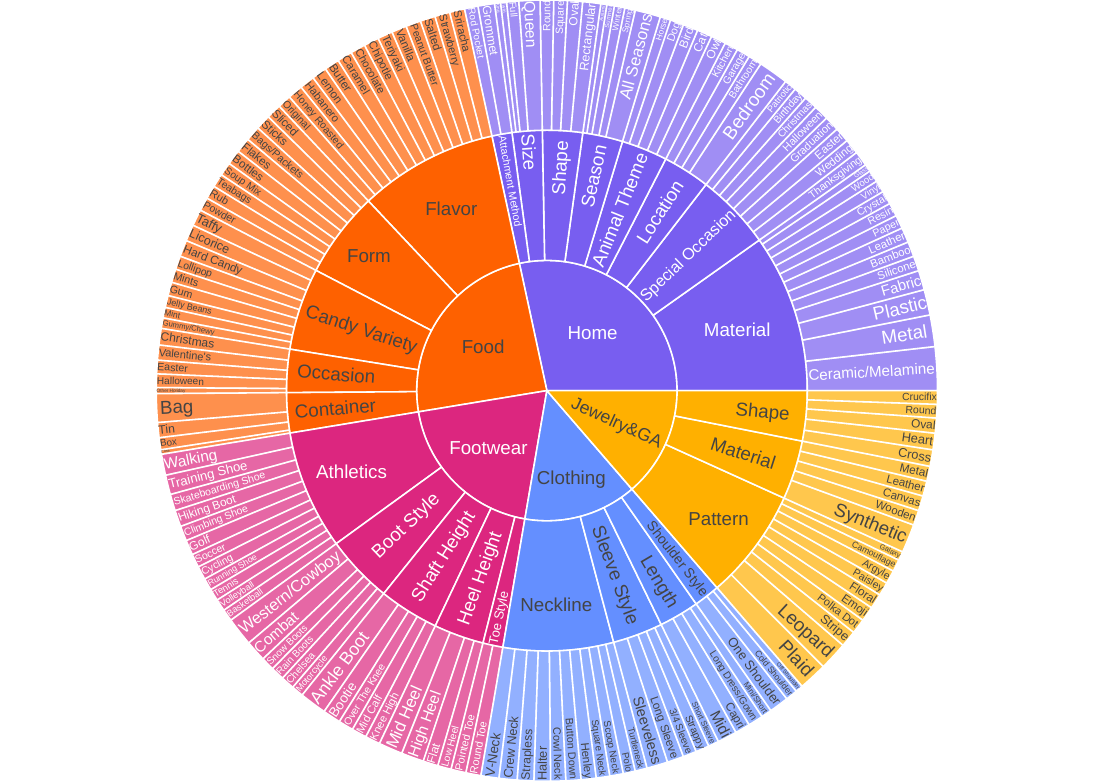}
         \caption{Evaluation Set}
         \label{fig:visual_data_distribution_evaluation}
     \end{subfigure}
        \caption{Data distribution of domains, attributes, and attribute values for training and evaluation sets. (A full-size version is attached to our appendix - Figure \ref{fig:appendix_large_visual_data_distribution})}
        \label{fig:visual_data_distribution}
\end{figure}

The overall domain-level dataset statistics is provided in Table \ref{tab:dataset_statistics_domain}.  We have 68,604 training instances and 1,610 high-quality evaluation instances. Our dataset covers 5 diverse domains and 25 carefully curated attributes specially for the task of implicit attribute value extraction. We also provide detailed attribute-level statistics in Table \ref{tab:dataset_statistics_attribute_level}. Different attributes contain different numbers of value options that are meticulously selected and processed and we have a total of 158 diverse attribute values. In addition, we visualize the data distribution of domains, attributes and their values for our training set and evaluation set in Figure \ref{fig:visual_data_distribution}(a) and \ref{fig:visual_data_distribution}(b), respectively. It can be observed that compared to the training set, each attribute in the evaluation set has a considerably balanced value distribution, making it more suitable for zero-shot MLLMs evaluation.


\begin{table*}[!thb]
\centering
\resizebox{\textwidth}{!}{%
\begin{tabular}{@{}lccccccc@{}}
\toprule
\textbf{Method} & \textbf{Language Model} & \textbf{Clothing} & \textbf{Footwear} & \textbf{Jewelry\&GA} & \textbf{Food} & \textbf{Home Product} & \textbf{All} \\ \midrule
\textit{Zero-shot methods} & \multicolumn{1}{l}{} & \multicolumn{1}{l}{} & \multicolumn{1}{l}{} & \multicolumn{1}{l}{} & \multicolumn{1}{l}{} & \multicolumn{1}{l}{} & \multicolumn{1}{l}{} \\
BLIP-2 & FlanT5XL-3B & 38.05 & 	49.21 & 	72.72 & 	61.54 & 	70.02 & 	59.75\\
BLIP-2 & FlanT5XXL-11B &55.31 & 	55.21 & 	82.72 & 	71.02 & 	71.33 & 	67.39\\
InstructBLIP & Vicuna-7B & 47.79 & 	48.26 & 	76.81 & 	61.28 & 	63.02 & 	59.43 \\
InstructBLIP & FlanT5XXL-11B & \textbf{62.83} & 	63.41 & 	83.18 & 	73.58 & 	73.96 & 	71.49 \\
LLaVA & Vicuna-7B & 22.12 & 	39.74 & 	62.72 & 	49.23 & 	57.76 & 	47.82 \\
LLaVA-1.5 & Vicuna-7B &26.54 & 	\textbf{67.72 }& 	41.95 & 	73.85 & 	66.96 & 	59.69 \\
LLaVA-1.5 & Vicuna-13B & 49.12 & 	63.72 & 	81.81 & 	76.15 & 	\textbf{80.31} & 	\textbf{71.86}\\
Qwen-VL-Chat & Qwen-7B & 32.30 & 	41.01 & 	67.27 & 	55.64 & 	57.11 & 	51.49 \\
Qwen-VL & Qwen-7B & 59.73 & 	57.72 & 	\textbf{84.09} & 	\textbf{76.92} & 	73.96 & 	70.86 \\
\midrule
GPT-4V & - & \textcolor{blue}{\textbf{77.43}} & 	\textcolor{blue}{\textbf{81.39}} & 	\textcolor{blue}{\textbf{90.45}} & 	\textcolor{blue}{\textbf{90.77}} & 	\textcolor{blue}{\textbf{89.93}} & 	\textcolor{blue}{\textbf{86.77}} \\
\midrule
\textit{Representative \& SOTA methods (Fine-Tuned)} & \multicolumn{1}{l}{} & \multicolumn{1}{l}{} & \multicolumn{1}{l}{} & \multicolumn{1}{l}{} & \multicolumn{1}{l}{} & \multicolumn{1}{l}{} & \multicolumn{1}{l}{} \\
DEFLATE (ACL 2023) & T5-Base-770M & 54.42 & 71.61 & 67.73 & 52.56 & 61.71 & 61.24 \\
LAVIN (NeurIPS 2023) & LLaMA-7B & \textbf{65.93} & \textbf{75.39} & \textbf{78.64} & \textbf{60.77} & \textbf{64.33} & \textbf{67.83} \\ \bottomrule
\end{tabular}%
}
\caption{Domain-level results. Analysis and representative error cases are provided in Section \ref{sec:analysis_domain}. \textbf{Bold black} shows best results in each block (zero-shot or finetuning), \textcolor{blue}{\textbf{bold blue}} shows best results overall.} 
\label{tab:result_domain_level}
\vspace{-4mm}
\end{table*}

\section{Experiment \& Benchmark}

In this section, we describe our experiment results evaluating the effectiveness of various MLLMs and the previous SOTA method on our ImplicitAVE dataset in diverse settings. 

\subsection{Experimental Setting}
\paragraph{Evaluation Setup}
We benchmark different models on our datasets from both attribute and domain levels: 

\noindent \textbf{$\bullet$ Attribute-Level Results} refer to the micro-F1 score calculated between the ground truth answer and the model-generated answer for \textit{each} query/interested \textit{attribute}. 

\noindent \textbf{$\bullet$ Domain-Level Results} refer to the micro-F1 score calculated between the ground truth answer and the model-generated answer for \textit{all} query/interested \textit{attributes} in \textit{each domain}.

\vspace{1mm}
\noindent We determine whether the generated answer is correct by checking whether the generated answer contains the true answer.

\paragraph{Models for Zero-Shot}
We utilize the following multimodal LLM frameworks in zero-shot settings:

\noindent \textbf{$\bullet$ BLIP-2}~\cite{BLIP-2} proposes a Query Transformer and employs an efficient two-stage vision-and-language pre-training strategy leveraging a frozen image encoder and an LLM. We provide benchmarks of BLIP-2 with two backbone LLM models, FLAN-T5-XL and FLAN-T5-XXL.

\noindent \textbf{$\bullet$ InstructBLIP}~\cite{Dai2023InstructBLIPTG} enhances vision-language models through instruction tuning with an instruction-aware Query Transformer introduced. We also report the performance with two backbone LLMs, Vicuna-7B and FLAN-T5-XXL.

\noindent \textbf{$\bullet$ LLaVA}~\cite{LLaVA} connects the visual encoder of CLIP~\cite{CLIP} with the language decoder, and performs fine-tuning on GPT-4 generated language-image instructions. We provide benchmarks of LLaVA with Vicuna-13B.

\noindent \textbf{$\bullet$ LLaVA-1.5}~\cite{Liu2023ImprovedBW} advances its predecessor by focusing on efficient visual instruction tuning, integrating a fully-connected vision-language cross-modal connector for enhanced interaction between visual and textual modality. We provide benchmarks of LLaVA-1.5 using Vicuna-7B and Vicuna-13B as the language models.

\noindent \textbf{$\bullet$ Qwen-VL}~\cite{bai2023qwen} proposes a novel visual receptor and a position-aware adapter, optimizing through a three-stage training pipeline on a multilingual and multimodal dataset. We report the performance of both Qwen-VL and the chat version, Qwen-VL-Chat.

\noindent \textbf{$\bullet$ GPT-4V}\footnote{https://chat.openai.com/} integrates vision into the GPT-4 architecture, one of the cutting-edge close-sourced LLMs fine-tuned by reinforcement learning from human feedback.

\begin{table*}[!thb]
\centering
\resizebox{\textwidth}{!}{%
\begin{tabular}{@{}llcccccccc@{}}
\toprule

{\textbf{Domains}} & {\textbf{Attributes}} & {\textbf{\# Values}} & InstructBLIP	& LLaVA 1.5	& Qwen-VL	& GPT-4V	& DEFLATE	& LAVIN\\
\textit{Language Model/Variants}  &  &  & \textit{FlanT5XXL-11B}	& \textit{Vicuna 13B}	& \textit{Qwen-7B} & \textit{-} & \textit{T5-Base-770M} & \textit{LLaMA-7B} \\

\midrule
Food	& Flavor	& 14	& 72.66	& 84.17	& 89.21	& \textcolor{blue}{\textbf{97.12}}	& 51.08	& 53.24\\
Home	& Material	& 13	& 74.05	& 61.39	& 67.09	& \textcolor{blue}{\textbf{84.81}}	& 77.22	& 82.28\\
Jewelry\&GA	& Pattern	& 10	& 81.08	& 80.18	& 89.19	& \textcolor{blue}{\textbf{90.99}}	& 61.26	& 78.38\\
Footwear	& Athletic Shoe Style	& 12	& 73.95	& 63.03	& 57.98	& \textcolor{blue}{\textbf{84.03}}	& 80.67	& 78.15\\
Clothing	& Neckline	& 11	& 53.64	& 25.45	& 52.73	& \textcolor{blue}{\textbf{78.18}}	& 50.91	& 57.27\\

\midrule
Food	& Form	& 9	& 70.93	& 59.30	& 75.58	& \textcolor{blue}{\textbf{86.05}}	& 67.44	& 81.40\\
Home	& Special Occasion	& 8	& 90.79	& 92.11	& 88.15	& \textcolor{blue}{\textbf{98.68}}	& 72.37	& 68.42\\
Clothing	& Sleeve Style	& 5	& 62.00	& 46.00	& 66.00	& 66.00	& 34.00	& \textcolor{blue}{\textbf{70.00}}\\
Footwear	& Boot Style	& 6	& 76.47	& 73.53	& 72.05	& \textcolor{blue}{\textbf{88.24}}	& 75.00	& 83.82\\
Jewelry\&GA	& Material	& 5	& 81.36	& 93.22	& 88.14	& \textcolor{blue}{\textbf{94.92}}	& 77.97	& 86.44\\

\midrule
Food	& Container	& 4	& 87.50	& \textcolor{blue}{\textbf{95.00}}	& 80.00	& 87.50	& 52.50	& 60.00\\
Footwear	& Heel Height	& 4	& 58.00	& 54.00	& 54.00	& \textcolor{blue}{\textbf{86.00}}	& 62.00	& 72.00\\
Clothing	& Shoulder Style	& 3	& \textcolor{blue}{\textbf{88.46}}	& 42.31	& 80.77	& 80.77	& 69.23	& 61.54\\
Home	& Attachment Method	& 2	& 45.00	& 100.00	& 100.00	& \textcolor{blue}{\textbf{100.00}}	& 90.00	& 90.00\\

\bottomrule
\end{tabular}%
}
\caption{Attribute-level results. Analysis and representative error cases are provided in Section \ref{sec:analysis_attribute}. Best results per attribute are shown in \textcolor{blue}{\textbf{bold blue}}.} 
\label{tab:result_attribute_level}
\end{table*}

\paragraph{Models for Finetuning} Due to the resource constraints, we fine-tuned and evaluated the following two open-source models in both few-shot and full-data tuning settings: 

\noindent \textbf{$\bullet$ LaVIN}~\cite{luo2023cheap} introduces a novel mix-of-modularity adaptation module, optimizing the integration of visual inputs into large language models through lightweight adapters and enabling efficient end-to-end training. 

\noindent \textbf{$\bullet$ DEFLATE}~\cite{zhang-etal-2023-pay} is a multi-modal generative-discriminative framework designed for both explicit and implicit attribute value extraction and is the previous SOTA model for implicit AVE.

\subsection{Experimental Results}
\label{sec:experiment_results}

\subsubsection{Domain-Level Results}
\label{sec:analysis_domain}
We present the domain-level results of all evaluated models in Table~\ref{tab:result_domain_level}. GPT-4V outperformed every other model in both the zero-shot and finetune setting in every single domain. Among the two models that were finetuned, LAVIN outperformed DEFLATE in every single domain by a minimum of 2.62 points (in the Home Product domain) and a maximum of 11.51 (in the Clothing domain). Among the open-source MLLMs, no single model outperformed all other models across all the domains, but Qwen-VL had the best scores in the Jewelry\&GA and Food domains. From Table~\ref{tab:result_domain_level} we also note that other than for LLaVA 1.5 in the Footwear domain, all other models that had multiple variants with different LLM sizes had significantly better performance on average from the variant with the larger size LLM in each domain in comparison to the variant with the smaller sized LLM. For example, in the Clothing domain, there was a minimum improvement of 15.04 micro-F1 points from the model variant with the smaller LLM (InstructBLIP w/ Vicuna 7B) to the model variant with the larger LLM (InstructBLIP w/ FLAN-T5-XXL) and an overall average of 18.29 micro-F1 point increase when using a model variant with a larger LLM in the Clothing domain. Similar trends can be seen among all domains.  


Additionally, among zero-shot methods, Clothing had the lowest micro-F1 across all domains for all models and model variants except for BLIP2 w/ FLAN-T5-XL and Qwen-VL. This leads us to believe that the Clothing domain is the most challenging domain in the dataset. We performed a comprehensive manual investigation and we believe there are two \textit{\underline{primary reasons} why the Clothing domain presents more challenges}, while other domains such as the Home domain are comparatively easier (We show examples from our manual investigation in Figures \ref{fig:error_analysis_clothing}, \ref{fig:error_analysis_home} for clarity): 

\paragraph{(1) Attributes within the Clothing domain demand a more nuanced understanding of local details in product images.} For example, the attribute ‘Sleeve Style’ in cases 1-4 and ‘Neckline’ in cases 7-12 (Figure \ref{fig:error_analysis_clothing}). In contrast, attributes in the home domain only require a global understanding of product pictures and text, such as attribute ‘Special Occasion’ in cases 13-16, ‘Shape’ and ‘Material’ in cases 17 and 21 (Figure \ref{fig:error_analysis_home}).

\paragraph{(2) The values of attributes in the Home domain are significantly more straightforward to identify compared to those in the Clothing domain.} For instance, the attribute `Special Occasion' includes values like [`Birthday', `Christmas', `Easter', `Graduation', `Halloween', `Patriotic', `Thanksgiving'], which are clearly more distinguishable than the values for 'Sleeve Style' [`Sleeveless', `Long Sleeve', `3/4 Sleeve', `Strappy', `Short Sleeve'] in the Clothing Domain.




\subsubsection{Attribute-Level Results}
\label{sec:analysis_attribute}
Table~\ref{tab:result_attribute_level} presents the attribute-level performance of all evaluated models.
As was observed in Table~\ref{tab:result_domain_level}, GPT-4V vastly outperforms all other models. We can see in Table~\ref{tab:result_attribute_level} that only in the `Shoulder Style' (InstructBLIP), `Container' (LLaVA 1.5) and `Sleeve Style' (LAVIN) attributes do a model outperform GPT-4V. InstructBLIP struggled significantly with the `Attachment Method' attribute as did LLaVA 1.5 with `Shoulder Style' compared to other models. 
On the other hand, Table~\ref{tab:result_attribute_level} shows that both finetuned models perform better than all of the open-source MLLMs in the zero-shot setting for the `Heel Height' attribute. This may indicate that there are attributes within the dataset for which prior pretrained knowledge of MLLMs is not sufficient for implicit value extraction of these attribute values and finetuning is needed to learn the mapping between instances of these attributes and the correct attribute values belonging to them. 

In addition, all models struggled on the `Sleeve Style' and `Neckline' attributes compared to each model's performance on other attributes. Representative error cases for different attributes are presented in Figures \ref{fig:error_analysis_clothing} and \ref{fig:error_analysis_home} in Appendix \ref{sec:appendix_error_analysis_comprehensive} along with a comprehensive error analysis. Here we provide our observations from the \textit{\underline{attribute-level error analysis}}:


\vspace{0.18cm}
\label{line:error_analysis_attribute}
\noindent \textbf{(1) Models often confuse attribute values that are similar yet distinct}, such as `3/4 Sleeve' versus `Long Sleeve' in cases 1-2, `Short Sleeve' versus `Sleeveless' in cases 3-4, and `Crew Neck' versus `Scoop Neck' in case 8 (Figure \ref{fig:error_analysis_clothing}).

\vspace{0.18cm}
\noindent \textbf{(2) Attributes that demand a detailed understanding of small image parts typically challenge models}, leading to errors. For instance, mistakes in identifying `Shoulder Style’ in cases 5-6 and `Neckline’ in cases 7-9 (Figure \ref{fig:error_analysis_clothing}).

\vspace{0.18cm}
\noindent \textbf{(3) Errors can also arise from conflicting modality inferences}, as seen in case 13 (Figure \ref{fig:error_analysis_home}), where the word `Snow Village' in the product text suggested Christmas, but the image aligned more with Halloween.



\begin{table}[!tb]
\centering
\resizebox{\columnwidth}{!}{%
\begin{tabular}{@{}lccccc@{}}
\toprule
\textbf{Domains} & GPT-4V & Qwen-VL & LLaVA-1.5 & InstructBLIP & BLIP-2 \\ \midrule
Clothing & 77.43 & 59.73 & 49.12 & 62.83 & 55.31 \\
Footwear & 81.39 & 57.72 & 67.72 & 63.41 & 55.21 \\ \midrule
\textbf{Attributes} &  &  &  &  &  \\ \midrule
Sleeve Style & 66.00 & 66.00 & 46.00 & 62.00 & 50.00 \\
Shaft Height & 63.33 & 35.00 & 61.66 & 26.67 & 30.00 \\
Season & 65.00 & 57.50 & 65.00 & 60.00 & 62.50 \\
Neckline & 78.18 & 52.73 & 25.45 & 53.64 & 48.18 \\ \midrule
\textbf{Average} & \textbf{68.13} & {\ul 52.81} & 49.53 & 50.58 & 47.67 \\ \bottomrule
\end{tabular}%
}
\caption{Examples of challenging domains \& attributes.}
\label{tab:result_challenging_domains_attributes}
\vspace{-5mm}
\end{table}

\subsubsection{Challenges and Opportunities}
\label{sec:challenges_opportunities}
\textbf{Challenging Domains \& Attributes:} It can be observed in Table \ref{tab:result_domain_level}, \ref{tab:result_attribute_level} that GPT-4V works well on some domains and attributes, but not on all of them, e.g., it only achieves 77.4\% micro-F1 on the Clothing domain and 66.0\% for the Sleeve Style attribute. Some examples of challenging domains, attributes, and the performance of various MLLMs are highlighted in Table \ref{tab:result_challenging_domains_attributes}. Besides, we can observe that the open-source models are lagging behind GPT-4V in many domains and attributes, and our dataset provides a good benchmark that points out the gap between them and provides opportunities for researchers to close it.

Furthermore, inspired by the error cases in Section \ref{line:error_analysis_attribute} and Appendix \ref{sec:appendix_error_analysis_comprehensive}, we point out some \textit{\underline{remaining challenges and opportunities}}:

\paragraph{\textbf{Model-Aspect:}}
(1) Enhance the ability to understand image details, including small areas and text in images; (2) Devise mechanisms to distinguish similar attribute values; (3) Properly handle conflicting modality inferences; (4) Reduce the performance gap in implicit AVE between open-source models and advanced closed models like GPT-4V.
\paragraph{\textbf{Dataset-Aspect:}}
Our ImplicitAVE dataset does not consider multi-valued attributes and negative instances, i.e., "none" as attribute values. We leave this extension for future work.

\begin{figure}[!tbh]
\centering
 \includegraphics[width=\columnwidth]{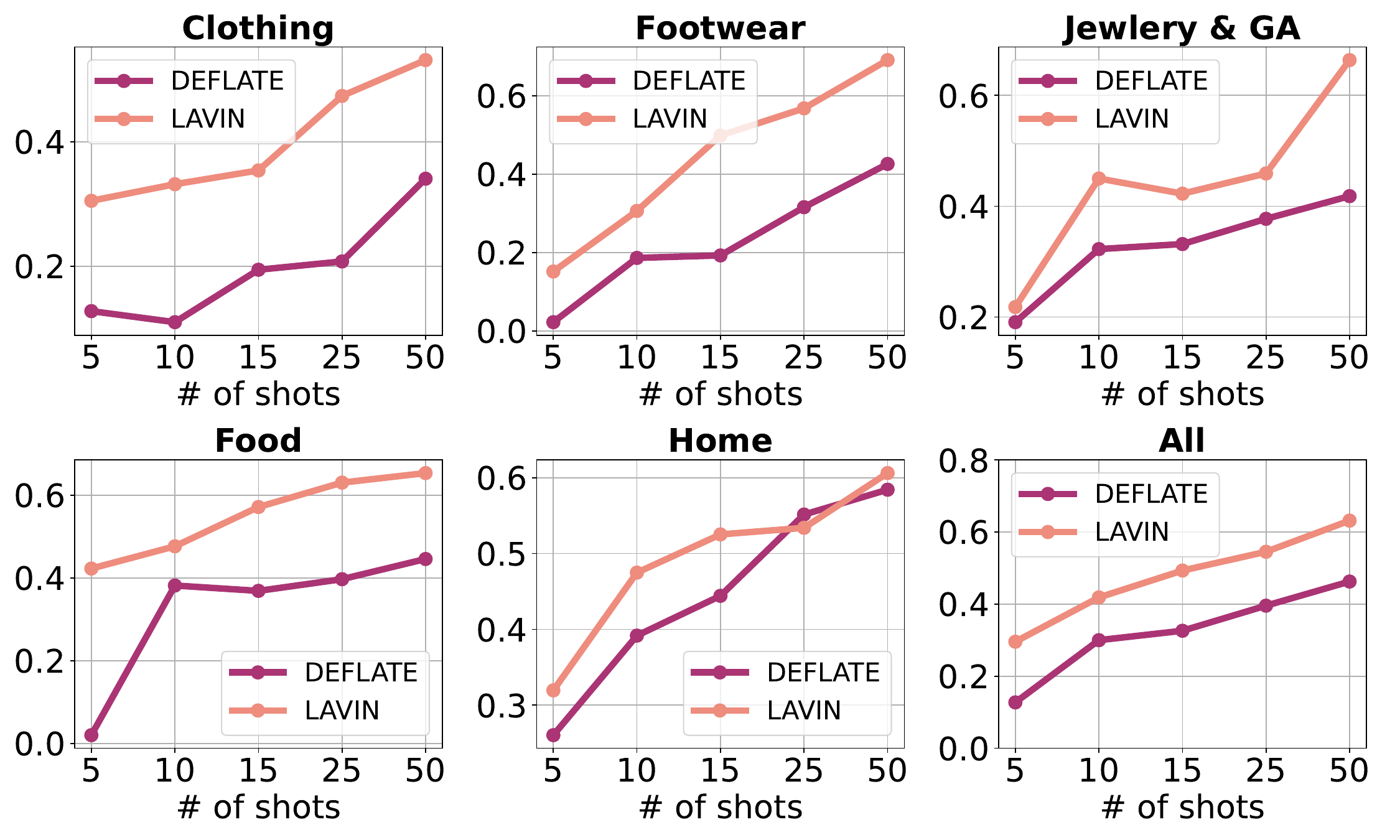}
 \caption{Performance comparison in few-shot settings of different domains.}
 \label{fig:com_fewshot}
 \vspace{-3mm}
\end{figure}

\begin{figure}[!tbh]
     \centering
     \begin{subfigure}[b]{\columnwidth}
         \centering
         \includegraphics[width=\textwidth]{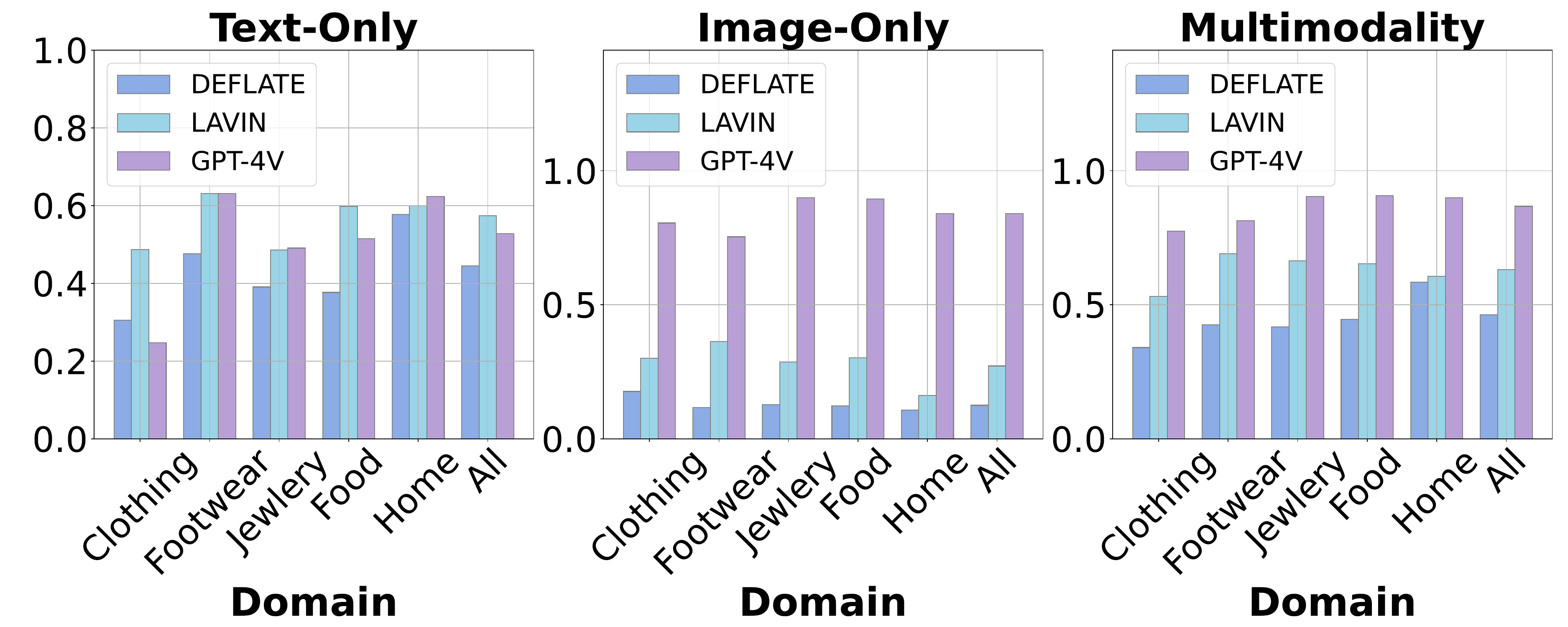}
         \caption{Comparison of methods.}
         \vspace{1pt}
         \label{fig:com_modal}
     \end{subfigure}
     \hfill
     \begin{subfigure}[b]{\columnwidth}
         \centering
         \includegraphics[width=\textwidth]{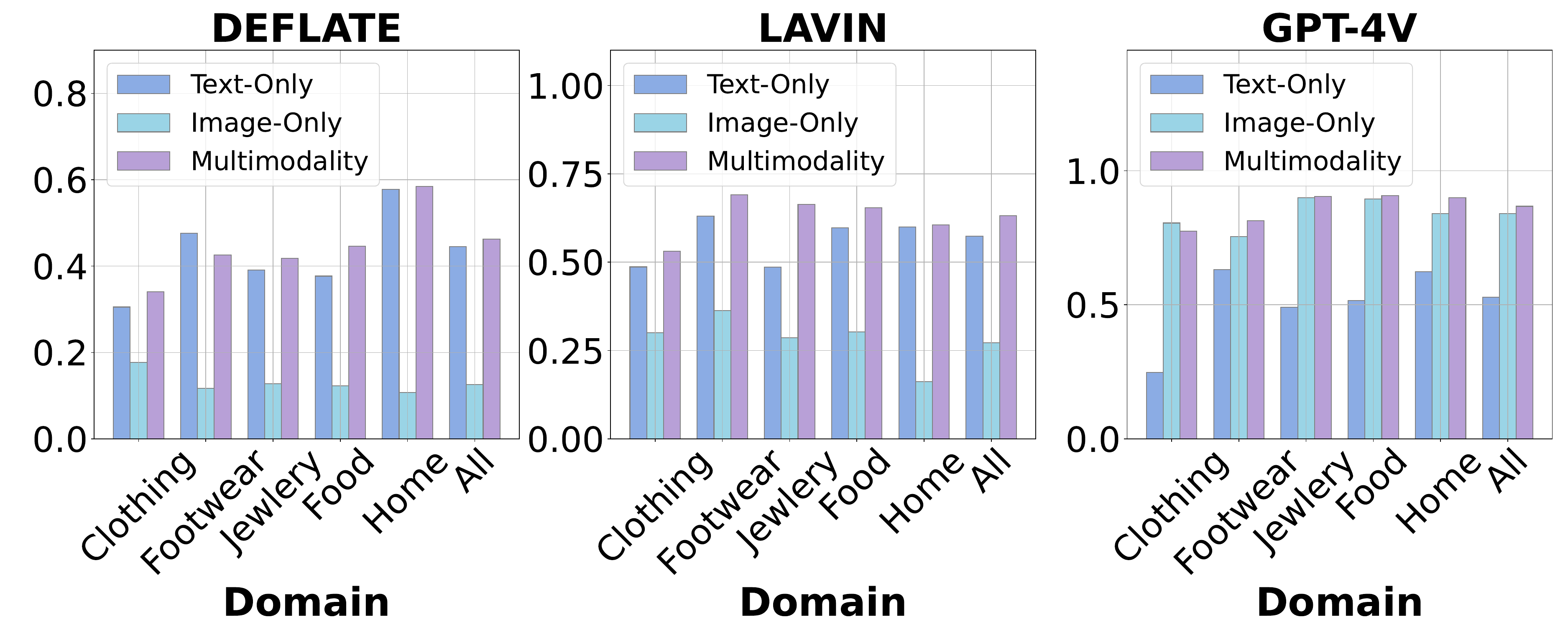}
         \caption{Comparison of modality.}
         \label{fig:com_method}
     \end{subfigure}
        \caption{Performance comparison of DEFLATE, LAVIN, and GPT-4V on different modalities.}
        \label{fig:com_method_modality}
        \vspace{-3mm}
\end{figure}


\subsubsection{Few-Shot Results}
Figure~\ref{fig:com_fewshot} shows the performance comparison of DEFLATE and LAVIN models in various few-shot settings. We note that in most K-shot settings, LAVIN outperforms DEFLATE by a noticeable amount. 
Also, we notice that different domains performed differently for the two models. In the 5-shot setting, `Food' was the lowest scoring domain for DEFLATE whereas `Food' was the highest scoring domain for LAVIN, but for 10-shot the domain trends for both models became similar (i.e., `Food' and `Home Product' domains were the two best-performing domains and `Clothing' and `Footwear' were the worst performing domains). For DEFLATE from 25-shot to 50-shot, the largest increase in micro-F1 was for the `Clothing' and `Footwear' domains whereas the increase was less significant for the other domains. This indicates that the model's ability to learn the attributes and attribute values in the `Clothing' and `Footwear' domains may continue to increase as the number of training examples increases. On the contrary, LAVIN saw the biggest increase in micro-F1 for the `Jewelry\&GA' and `Footwear' domains thereby hinting that increasing the training examples for these domains in LAVIN would enable the model to substantially increase its ability to categorize instances of these two domains. 

\subsubsection{Modality-Level Results}
Figure~\ref{fig:com_method_modality} visualizes performance comparisons of DEFLATE, LAVIN, and GPT-4V with different modalities.
Firstly, it is evident that for LAVIN and DEFLATE, the image-only modality performed extremely poorly compared to the text-only and combined modalities. This leads us to believe that these models’ image understanding capabilities may be too poor to extract implicit value from product images. However, it is worth noting that in all domains except `Footwear' for DEFLATE, both LAVIN and DEFLATE perform better in the multimodal modality over the text-only modality thereby indicating that the image information does in fact help the model predict attribute values of instances. With GPT-4V we notice a very high performance in the image-only modality and only minimal improvement with the multimodal modality in comparison to the image-only modality. This speaks to the strength of GPT-4V’s zero-shot image classification capabilities, especially in comparison to LAVIN and DEFLATE. Even though GPT-4V boasts impressive performance in most regards, it is worth noting that GPT-4V’s text-only modality performance in the `Clothing' domain was especially poor. It scored even lower than the text-only scores of LAVIN and DEFLATE and, in the `Clothing' domain, the multimodal performance for GPT-4V was lower than the image-only performance thereby indicating that the text component confused the model causing it to perform worse than it did without the text component.

\begin{figure}[!tbh]
\centering
 \includegraphics[width=\columnwidth]{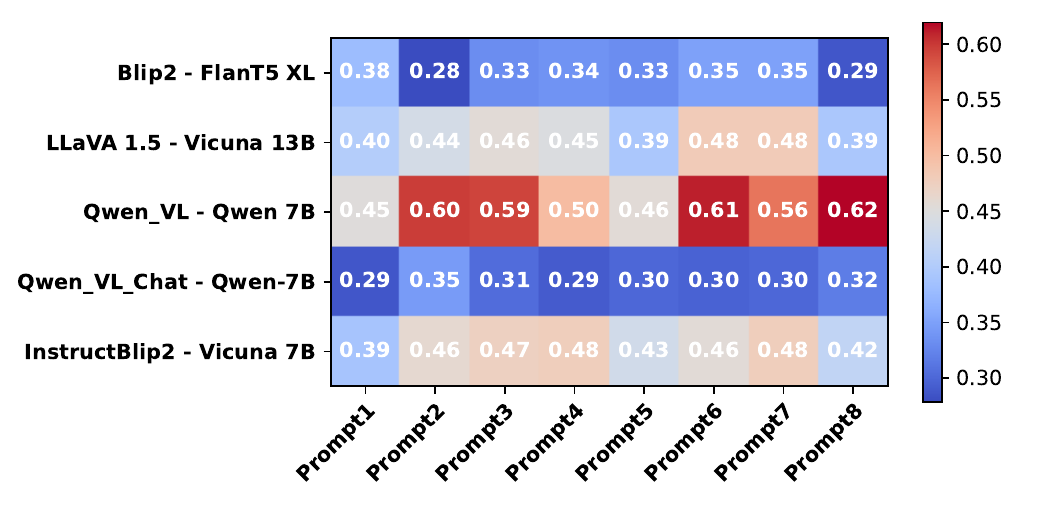}
 \caption{The influence of prompts (detailed in Table~\ref{tab:prompt} in Appendix~\ref{sec:prompt}) on different models.}
 \label{fig:prompts}
 \vspace{-3mm}
\end{figure}

\subsection{Ablation Study on Prompt Templates}
In order to obtain baseline results that accurately reflect the quality of our dataset we conducted ablations on the prompt for the open-source MLLMs. Observing drastic micro-F1 score differences on the evaluation set by using different prompts in the early stages of experimenting led us to conduct a standardized ablation study on 8 different prompts listed in Table \ref{tab:prompt}. Each prompt had three components: context containing the title of the product with explicit mention of the attribute value removed, question, and options to answer from. To conduct a fair evaluation of the prompts, across all models we fixed the random seed at 42 as well as the hyperparameters: temperature = 1, top\_p = 0.8, max\_new\_tokens = 17, min\_length = 1, and num\_beams = 5. Our results are shown in Fig~\ref{fig:prompts} and the best prompt for each model type was used for all variants of that model.

\section{Related Work}
\subsection{Attribute Value Extraction Dataset}
Attribute Value Extraction (AVE) has emerged as a crucial task for online shopping, aiming to identify the values of product attributes from various data sources. 
At the heart of many e-commerce applications, such as product comparison, retrieval, recommendation, and the construction of product graphs and online shop assistants, lies the extraction of attribute values \cite{kddtutorial}. 
Although several AVE datasets have been introduced, each exhibits certain limitations, as shown in Table \ref{tab:dataset_compare}.

The OpenTag dataset \cite{opentag}, one of the early datasets collected from Amazon, highlights the importance of open-world value sets. In contrast, the AE-110K dataset \cite{xu-etal-2019-scaling} expands the scope of AVE datasets to include more products, a broader range of attributes, and denser attribute coverage per product, though it lacks human expert annotation. The AdaTag dataset \cite{yan-etal-2021-adatag} focuses on the rich information contained in product bullets, excluding product descriptions, which facilitates more efficient training and inference for such tasks but lacks diversity in product domains and is not publicly available. The MAVE dataset \cite{mave}, a large public dataset for AVE research, encompasses a wide range of categories and diverse attributes, constructing structured product files as text inputs. However, in real-world scenarios, text information alone may not imply certain attributes of interest, making product images a complementary source of information for indicating or validating the answers to specific attributes. To address this, the MEPAVE \cite{zhu-etal-2020-multimodal} and DESIRE \cite{zhang-etal-2023-pay} datasets were introduced to include multimodal product information such as product titles, descriptions and images. While several explicit AVE datasets exist, implicit AVE is much more challenging and under-explored.  To advance multi-modal AVE research further, we introduce the first publicly available multimodal implicit AVE dataset, ImplicitAVE, featuring careful human annotation and a versatile range of items from multiple domains. Our dataset is considerably different from DESIRE, as detailed in Appendix \ref{sec:dif_desire}.



\subsection{Multimodal Large Language Models}
Multimodal Large Language Models (MLLMs) have demonstrated impressive performance on a variety of tasks \cite{BLIP-2, LLaVA,Liu2023ImprovedBW,bai2023qwen,ye2023mplug,luo2023cheap,dong2023musechat}.
BLIP-2 \cite{BLIP-2} uses frozen pre-trained image models and language models, and proposes a lightweight querying transformer Q-Former to bridge the two modalities. InstructBLIP \cite{Dai2023InstructBLIPTG} outperforms BLIP-2 \cite{BLIP-2} by using vision-language instruction tuning, where the instruction tuning data is collected from publicly available datasets, by manually transforming them into instruction tuning format. To improve the diversity and in-depth reasoning in the instruction, LLaVa \cite{LLaVA} proposes to use language-only GPT-4 to construct multimodal language-image instruction tuning data. mPLUG-Owl \cite{ye2023mplug} and Qwen-VL \cite{bai2023qwen} propose novel training paradigms for LLMs.
However, since most popular open-source MLLMs are parameter-heavy, LAVIN \cite{luo2023cheap} proposes a novel and efficient solution for vision-language instruction tuning by adopting lightweight modules, i.e., adapters, to bridge the gap between LLMs and vision modules, which does not require expensive vision-language pretraining to align text and image embedding beforehand. Despite achieving significant progress, the performance of MLLMs on implicit AVE has not been well-studied. Recent work EIVEN \cite{zou-etal-2024-eiven} finetuned an efficient MLLM framework for implicit AVE but did not compare with exiting MLLMs in zero-/few-shot settings. Our work establishes the first comprehensive benchmark of multimodal LLMs for implicit AVE under diverse settings and reveals intriguing
insights and potential future research directions in Section \ref{sec:challenges_opportunities}.


\vspace{-1pt}
\section{Conclusion}
\vspace{-2pt}

In this paper, we introduced ImplicitAVE, the first publicly accessible multimodal dataset specifically designed for implicit attribute value extraction, aimed at overcoming the limitations of existing datasets focused on explicit attribute values. By carefully curating attribute values and incorporating both implicit attribute values and product images, ImplicitAVE comprises 6.8K training instances and 1.6K human re-annotated high-quality evaluation instances across five diverse domains. Moreover, we benchmarked the performance of six recent multimodal large language models on it under diverse settings, highlighting the challenges of implicit value extraction. In the future, we plan to further expand our ImplicitAVE dataset to include multi-valued attributes and negative instances.

\section*{Acknowledgements}

This research is partially supported by NSF grant \#210751 and UIC DPI Seed Program. Any opinions, findings, and conclusions expressed here are those of the authors and do not necessarily reflect the views of NSF. We thank our reviewers for their insightful feedback and comments which helped improve the quality of our paper. 

\section{Limitation}
Our ImplicitAVE dataset does not consider multi-valued attributes and negative instances, i.e. "none" as attribute values. We leave this extension as future work. Due to computational resource constraints and limited budgets, we did not evaluate open MLLMs with parameters larger than 13B.

\section{Ethics Statement}

The datasets that we sourced from are publicly available. In this work, we propose a multimodal Implicit AVE dataset and provide a comprehensive benchmark of MLLMs. We do not expect any direct ethical concern from our work.

\bibliography{custom}

\clearpage
\appendix

\section{Novelty and Contribution of Dataset}
\label{sec:dif_desire}
In Table \ref{tab:dataset_compare}, we compare our dataset with existing AVE datasets from different aspects. While several \textbf{\textit{explicit}} AVE datasets exist, \textbf{\textit{implicit}} AVE is much more challenging and underexplored. To the best of our knowledge, our work introduces the first open-source dataset that is expressly designed to address the task of implicit AVE. Here, we would like to clarify that our dataset is \textbf{considerably different} from DESIRE \cite{zhang-etal-2023-pay} regarding:

\begin{enumerate}
  \item \textbf{Accessibility:} In DESIRE, all data are encrypted, and image data are encoded by DALL-E \cite{pmlr-v139-ramesh21a}, with raw images unavailable. Moreover, DESIRE does not provide statistics on how many implicit AVE examples are included. In contrast, our dataset contains 70k+ curated implicit AVE examples, which are immediately available with raw images provided.
  \item \textbf{Domain Scope:} DESIRE only contains the `Food' domain, while our dataset contains five domains, including more challenging domains such as `Clothing' and `Footwear'.
  \item \textbf{Language:} DESIRE is based on Chinese while our dataset is in English.
\end{enumerate}

\section{Domain, Attribute, and Value}
\label{sec:appendix_domain_attribute_value}
\begin{figure*}[!th]
    \centering
    \includegraphics[width=\textwidth]{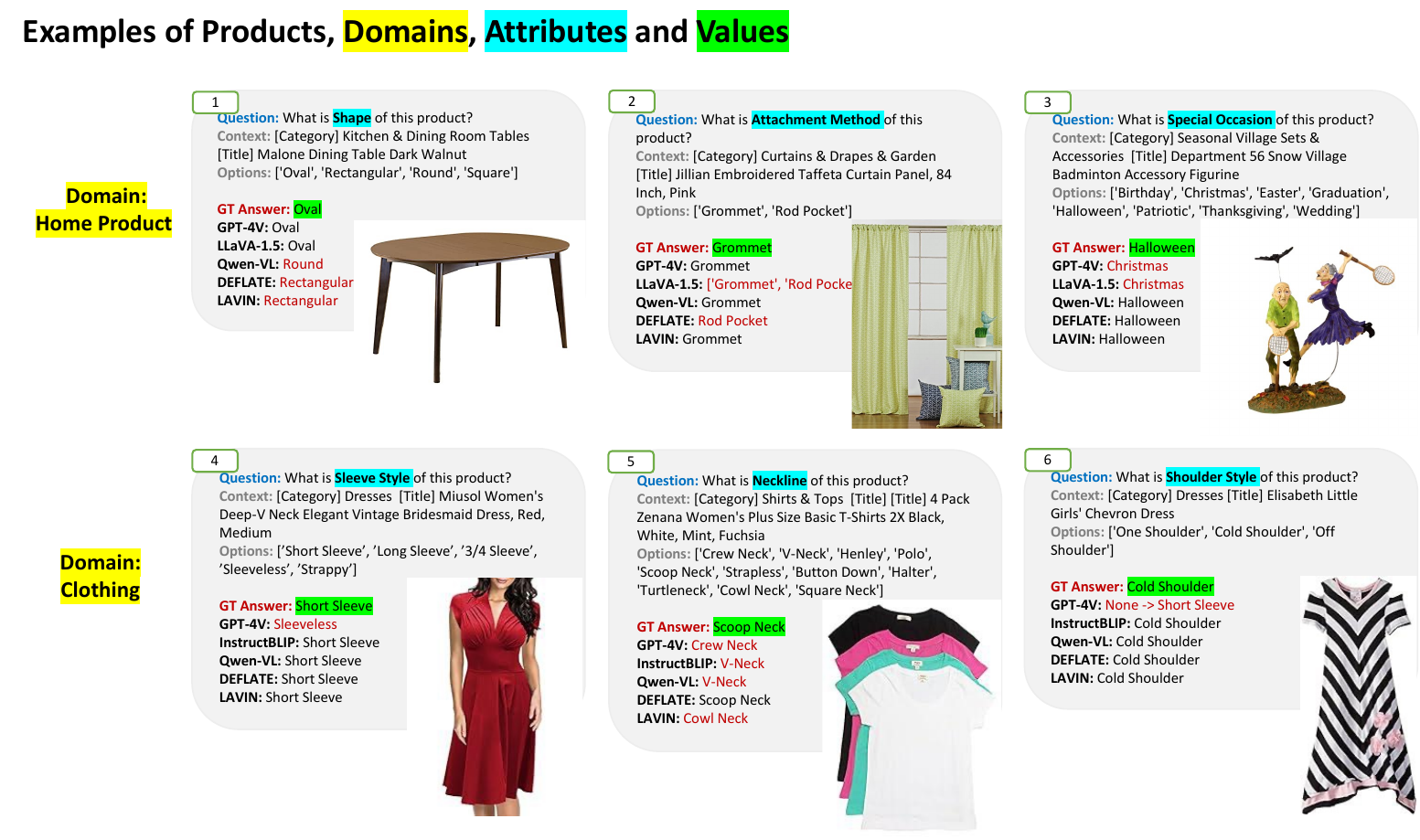} 
    \caption{Examples of products, domains, attributes, values.}
    \label{fig:domain_attribute_value}
\end{figure*}

The \textbf{ontology} of our data adheres to the \textbf{\textit{domain-attribute-value}} structure, where (1) Each domain contains relevant attributes that characterize different aspects of the domain product. For instance, the `Clothing' domain contains attributes such as `Sleeve Style' and `Neckline'; (2) Each attribute comprises a set of possible values (also called "attribute values") and we aim to extract its ground truth value from product images and text contexts. For example, the attribute `Sleeve Style' may include values such as `Long-sleeve', `3/4 sleeve', and `Strappy'. The full details are depicted in Table \ref{tab:dataset_statistics_attribute_level}. Figure \ref{fig:domain_attribute_value} also presents a few examples of products in different domains, with different attributes and values.

Therefore, in Tables \ref{tab:result_domain_level} and \ref{tab:result_attribute_level}, \textbf{Attribute-level results} refer to the micro-F1 score calculated between the ground truth answer and the model-generated answer for each query/interested attribute. \textbf{Domain-level results} refer to the micro-F1 score calculated between the ground truth answer and the model-generated answer for all query/interested attributes in each domain. We determine whether the generated answer is correct by checking whether the generated answer contains the true answer.

\section{Detailed Error Analysis and Remaining Challenges}
\label{sec:appendix_error_analysis_comprehensive}
\begin{figure*}[!th]
    \centering
    \includegraphics[width=\textwidth]{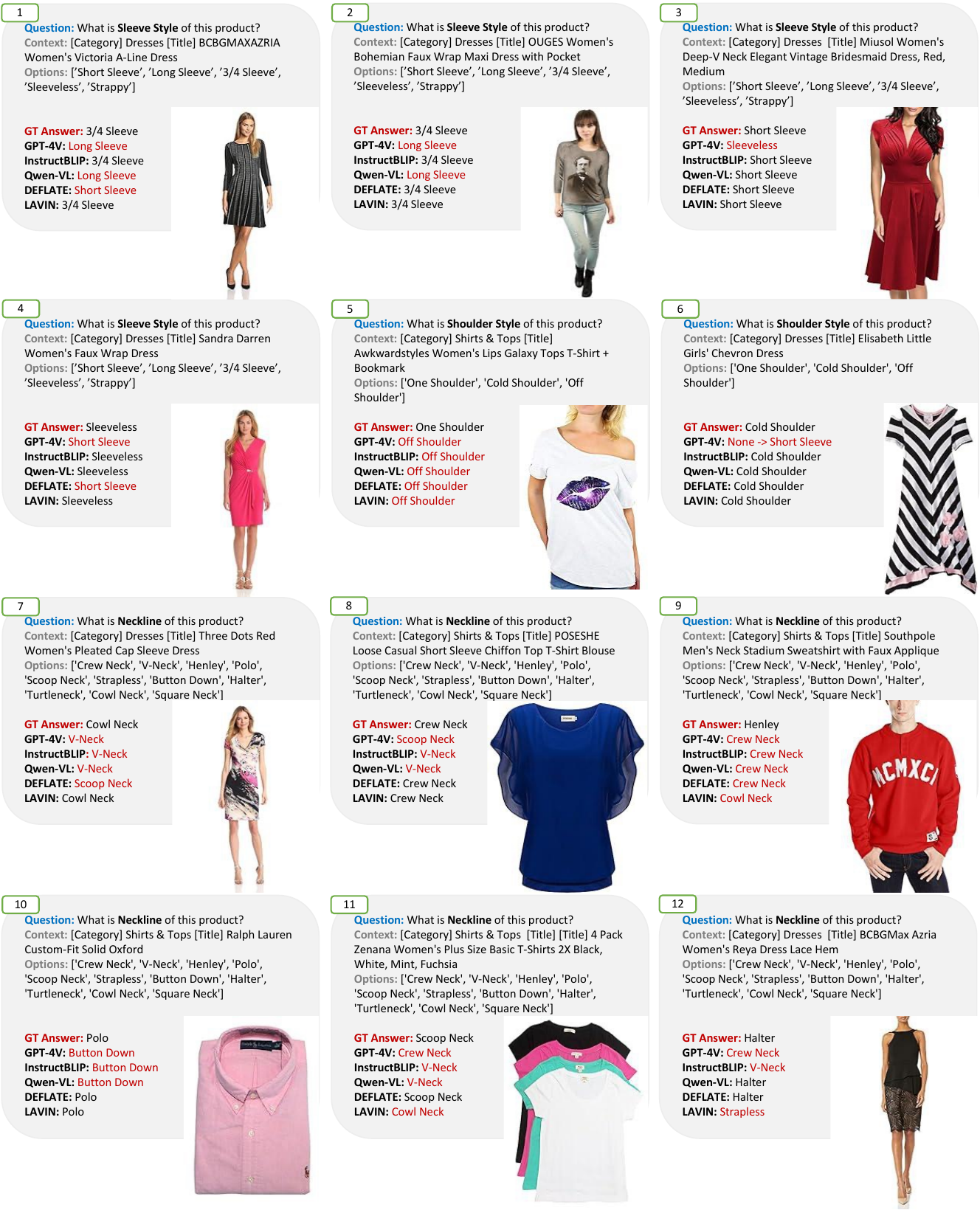} 
    \caption{Representative error cases - clothing domain. (Domain-level analysis: Section \ref{sec:analysis_domain}; Attribute-level analysis: Section \ref{sec:analysis_attribute}; Comprehensive error analysis and remaining challenges: Appendix \ref{sec:appendix_error_analysis_comprehensive})}
    \label{fig:error_analysis_clothing}
\end{figure*}

\begin{figure*}[!th]
    \centering
    \includegraphics[width=\textwidth]{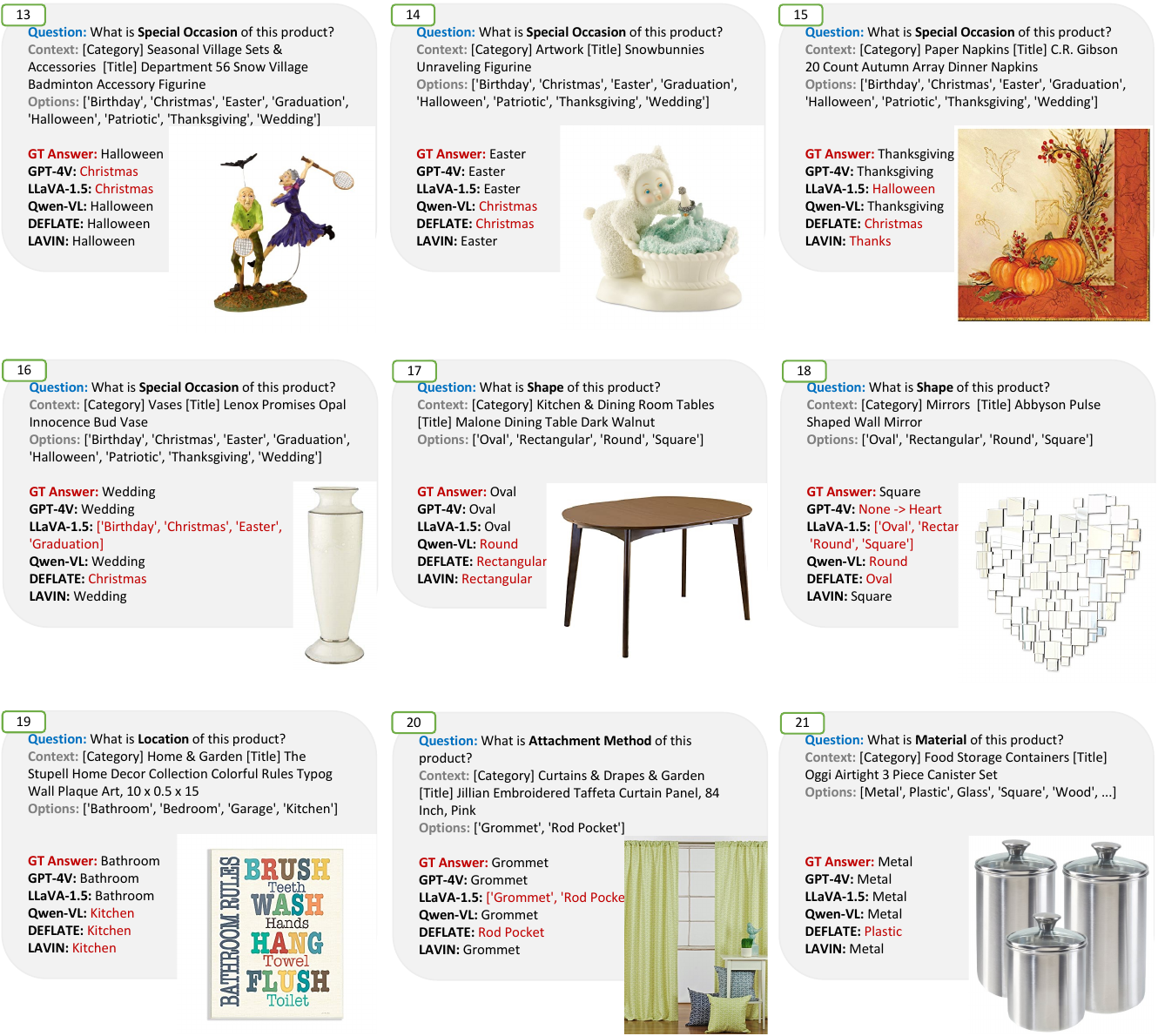} 
    \caption{Representative error cases - home domain. (Domain-level analysis: Section \ref{sec:analysis_domain}; Attribute-level analysis: Section \ref{sec:analysis_attribute}; Comprehensive error analysis and remaining challenges: Appendix \ref{sec:appendix_error_analysis_comprehensive})}
    \label{fig:error_analysis_home}
\end{figure*}

We have conducted an exhaustive analysis of cases incorrectly predicted by various models, with a particular focus on GPT-4V. \textbf{Representative error cases} for different domains and attributes are presented in Figure \ref{fig:error_analysis_clothing} and Figure \ref{fig:error_analysis_home}. Here we provide a more detailed error analysis from different perspectives:

\paragraph{\textbf{Attribute-Level:}}
(1) Models often confuse attribute values that are similar yet distinct, such as `3/4 Sleeve' versus `Long Sleeve' in cases 1-2, `Short Sleeve' versus `Sleeveless' in cases 3-4, and `Crew Neck' versus `Scoop Neck' in case 8.
(2) Attributes that demand a detailed understanding of small image parts typically challenge models, leading to errors. For instance, mistakes in identifying `Shoulder Style’ in cases 5-6 and `Neckline’ in cases 7-9.
(3) Errors can also arise from conflicting modality inferences, as seen in case 13, where the word `Snow Village' in the product text suggested Christmas, but the image aligned more with Halloween.

\paragraph{\textbf{Domain-Level:}}
(1) `Clothing' is the hardest domain for most models because it contains many attributes that require fine-grained understanding of product images. For example, ‘Sleeve Style’ in cases 1-4 and ‘Neckline’ in cases 7-12.
(2) While easier domains such as `Home' usually consist of attributes that only need a more global understanding of product image and text context, such as ‘Special Occasion’ in cases 13-16, ‘Shape’ and ‘Material’ in cases 17 and 21.

\paragraph{\textbf{Model-Level:}}
(1) Open-source models are not good at recognizing and leveraging text in images. Taking case 19 as an example, Qwen-VL, DEFLATE, and LAVIN fail to utilize the text words ‘BATHROOM, Teeth, Toilet’ in the image.
(2) Interestingly and uniquely, when GPT-4V considers none of the provided options suitable, it will answer `None' and then give an answer it feels is a better match, as shown in cases 6, 18.
(3) LLaVA 1.5 tends to provide multiple answers in ambiguous situations, as can be seen in cases 16, 18.

Inspired by the above error cases, we point out some \textbf{remaining challenges and opportunities}:

\paragraph{\textbf{Model-Aspect:}}
(1) Enhance the ability to understand image details, including small areas and text in images; (2) Devise mechanisms to distinguish similar attribute values; (3) Properly handle conflicting modality inferences; (4) Reduce the performance gap in implicit AVE between open-source models and advanced closed models like GPT-4V.
\paragraph{\textbf{Dataset-Aspect:}}
Our ImplicitAVE dataset does not consider multi-valued attributes and negative instances, i.e. ``none'' as attribute values. We leave this extension for future work.

\section{Implementation Details}
All open-source MLLMs are evaluated on a single A100 GPU. Due to RAM and Disk space constraints, all model pre-trained weights were loaded in at 4 or 8 bits using the bitsandbytes library for quantization. All model variants using the Vicuna-13B or Flan-T5-XXL are loaded in at 4 bits, and all other models are loaded in at 8 bits. Additionally, eight different prompts are tested, with the best performance reported for each open-source model (BLIP2, InstructBLIP, LLaVA, LLaVA 1.5, Qwen-VL). A list of valid attribute value options is provided when prompting the MLLMs. For all settings, we use micro-F1/accuracy as evaluation metrics. The prompt templates are available in the Appendix \ref{sec:prompt}.

\section{Prompt Templates}
\label{sec:prompt}

Table \ref{tab:prompt} provides our prompt templates for all zero-shot methods except GPT-4V. The best results are displayed. The prompt template we use for GPT-4V is: ``What is the \{attribute\_names\} of this product? Context: [Category] \{category\} \{texts\}. Choose the most appropriate one from the options: \{Options\}.''

\begin{table*}[!tbh]
\centering
\resizebox{\textwidth}{!}{%
\begin{tabular}{@{}ll@{}}
\toprule
 \multirow{2}{*}{Prompt 1} & "Question: What is \{attribute\_names\} of this product?\textbackslash nContext: [Category] \{category\} \{texts\}.\textbackslash n\\
 & You must only answer the question with exactly one of the following options \{options\}.\textbackslash nAnswer:" \\
\midrule
\multirow{2}{*}{Prompt 2} & "What is \{attribute\_names\} of this product?[Category] \{category\} \{texts\}.\\
& Answer with the option from the given choices directly: \{options\}.\textbackslash nAnswer:"\\
 \midrule
\multirow{2}{*}{Prompt 3} & "[Category] \{category\} \{texts\}. What is \{attribute\_names\} of this product? \\
& Answer with the option from the given choices directly: \{options\}.\textbackslash nAnswer:"\\
\midrule
\multirow{2}{*}{Prompt 4} & "[Category] \{category\} \{texts\}. What is \{attribute\_names\} of this product based on the given information and the given image? \\
& Answer with the option from the given choices directly: \{options\}.\textbackslash nAnswer:"\\
\midrule
\multirow{2}{*}{Prompt 5} & "[Category] \{category\} \{texts\}. Which one of \{options\} is the \{attribute\_names\} of this product? \\
& Answer with the option from the given choices directly.\textbackslash nAnswer:"\\
\midrule
\multirow{2}{*}{Prompt 6} & "\{texts\}. What is the \{attribute\_names\} of this product? \\
& Answer with the option from the given choices directly: \{options\}.\textbackslash nAnswer:"\\
\midrule
\multirow{2}{*}{Prompt 7} & "\{texts\}. Based on the description and the image, what is the \{attribute\_names\} of this product? \\
& Answer with the option from the given choices directly: \{options\}.\textbackslash nAnswer:"\\
\midrule
\multirow{2}{*}{Prompt 8} & "What is the \{attribute\_names\} of this product: \{texts\}? \\
& Answer with the option from the given choices directly: \{options\}.\textbackslash nAnswer:"\\

\bottomrule
\end{tabular}%
}
\caption{Prompt templates for all zero-shot methods except GPT-4V. The best prompt for each model type was used for all variants of that model. The prompt used for GPT-4V is provided in Appendix \ref{sec:prompt} and our code.}
\label{tab:prompt}
\end{table*}

\clearpage

\begin{figure*}[!tbh]
     \centering
     \begin{subfigure}[b]{0.965\textwidth}
         \centering
         \includegraphics[width=\textwidth]{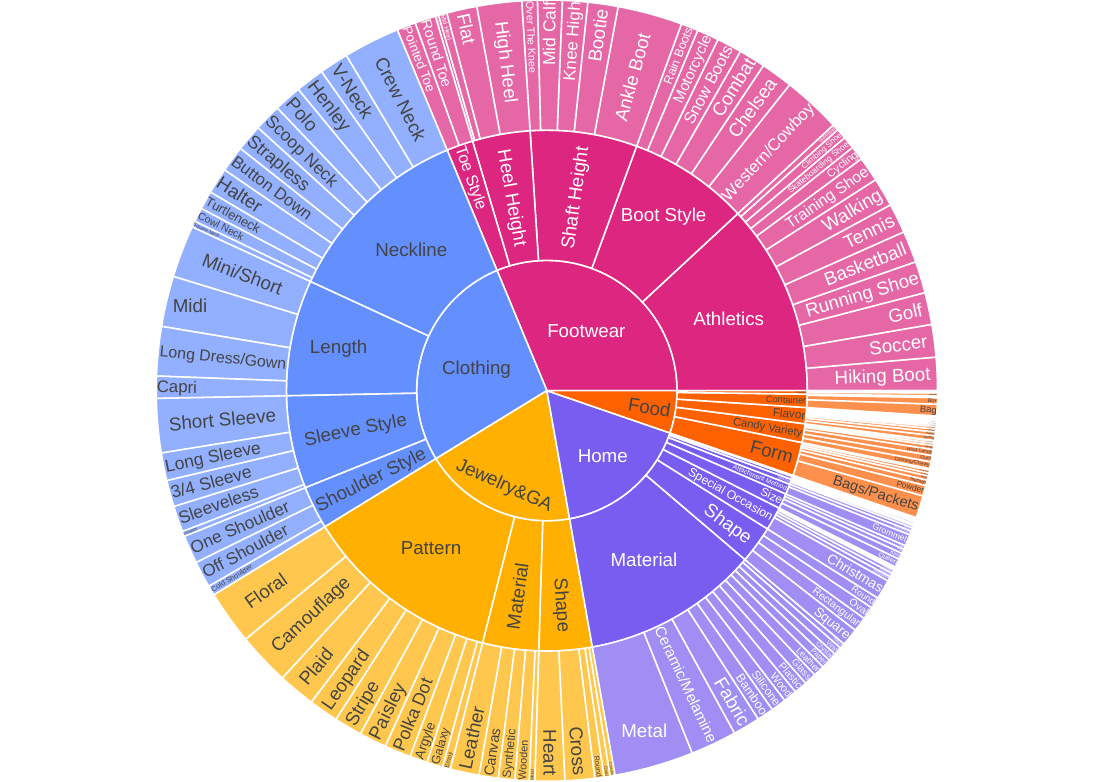}
         \caption{Training Set}
         \vspace{3pt}
     \end{subfigure}
     \hfill
     \begin{subfigure}[b]{0.965\textwidth}
         \centering
         \includegraphics[width=\textwidth]{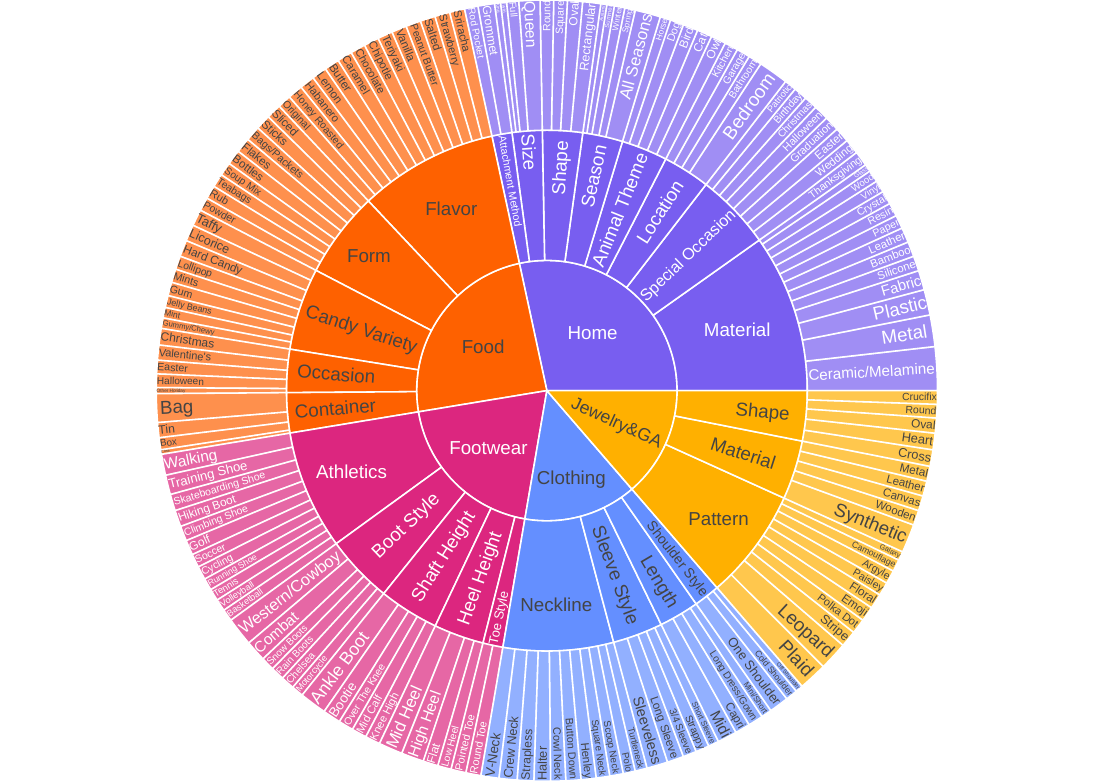}
         \caption{Evaluation Set}
         \vspace{3pt}
     \end{subfigure}
        \caption{Data distribution of domains, attributes, and attribute values for training and evaluation sets. The full-size version of Figure \ref{fig:visual_data_distribution}.}
        \label{fig:appendix_large_visual_data_distribution}
\end{figure*}

\end{document}